\documentclass[letterpaper, 10 pt, conference]{ieeeconf}  

\IEEEoverridecommandlockouts                              
\usepackage{amsmath} 
\usepackage{amssymb}  
\usepackage{graphicx} 
\usepackage{upgreek}
\usepackage{float}
\usepackage{bm}
\usepackage{colortbl}
\usepackage[dvipsnames]{xcolor}

\PassOptionsToPackage{svgnames}{xcolor}
\usepackage{tcolorbox}
\tcbuselibrary{skins,breakable}
\usetikzlibrary{shadings,shadows}

    {\endtcolorbox}

\title{\LARGE \bf
A Framework for a Capability-driven Evaluation of Scenario Understanding for Multimodal Large Language Models in \\ Autonomous Driving}

\author{Tin Stribor Sohn$^{1}$$^{*}$, Philipp Reis$^{2}$$^{*}$, Maximilian Dillitzer$^{1,3}$$^{*}$,\\Johannes Bach$^{1}$, Jason J. Corso$^{4}$ and Eric Sax$^{5}$
\thanks{*Equal Contribution}
\thanks{$^{1}$Dr. Ing. h.c. F. Porsche AG, Weissach, Germany
        {\tt\small tin\_stribor.sohn@porsche.de}}%
\thanks{$^{2}$Forschungszentrum Informatik, Karlsruhe, Germany
        {\tt\small reis@fzi.de}}%
\thanks{$^{3}$Hochschule Esslingen, Esslingen, Germany
        }%
\thanks{$^{4}$University of Michigan and Voxel51, Ann Arbor, USA
        }%
\thanks{$^{5}$Karlsruher Institut für Technologie, Karlsruhe, Germany
        }%
}

\begin{document}

\maketitle
\thispagestyle{empty}
\pagestyle{empty}

\begin{abstract}
Multimodal large language models (MLLMs) hold the potential to enhance autonomous driving by combining domain-independent world knowledge with context-specific language guidance. Their integration into autonomous driving systems shows promising results in isolated proof-of-concept applications, while their performance is evaluated on selective singular aspects of perception, reasoning, or planning. To leverage their full potential a systematic framework for evaluating MLLMs in the context of autonomous driving is required.
This paper proposes a holistic framework for a capability-driven evaluation of MLLMs in autonomous driving. The framework structures scenario understanding along the four core capability dimensions semantic, spatial, temporal, and physical. They are derived from the general requirements of autonomous driving systems, human driver cognition, and language-based reasoning. It further organises the domain into context layers, processing modalities, and downstream tasks such as language-based interaction and decision-making. To illustrate the framework’s applicability, two exemplary traffic scenarios are analysed, grounding the proposed dimensions in realistic driving situations. The framework provides a foundation for the structured evaluation of MLLMs’ potential for scenario understanding in autonomous driving.
\end{abstract}

\section{INTRODUCTION}

Multimodal large language models (MLLMs) demonstrate their capabilities across various vision-language tasks, including language-based information retrieval \cite{Sohn_2024}, visual question answering \cite{guo2023images}, image captioning \cite{chen2022visualgpt}, and vision-language navigation \cite{wang2020}.
Recently, their application in autonomous driving has gained increasing attention, where MLLMs enhance situational awareness \cite{Guo_2024} and decision-making \cite{Hwang_2024} by integrating multimodal sensory inputs with high-level reasoning.

Despite the promising potential of MLLMs for autonomous driving, current research predominantly presents proof-of-concept approaches addressing isolated aspects such as semantic scene description or reasoning-based behavioural planning. A comprehensive overview of the full spectrum of required capabilities for scenario understanding in autonomous driving is still lacking. Specifically, no unified framework exists to evaluate MLLMs holistically with regard to their ability to understand, describe, anticipate, and interact with dynamic traffic scenarios.

\begin{figure}[t]
    \centering
    \includegraphics[width=\linewidth]{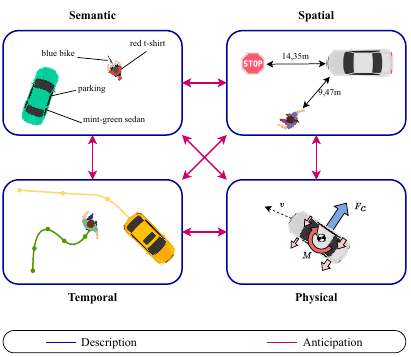}
    \caption{Four descriptive core capability dimensions of MLLMs form the basis for a capability-driven evaluation framework. The anticipation capability links all dimensions to form a holistic understanding relevant to the driving task in traffic scenarios.}
    \label{fig:Teaser}
\end{figure}

To address this gap, this paper proposes a systematic framework for the capability-driven evaluation of MLLMs in autonomous driving. The framework derives key capabilities from the scattered literature, as shown in Figure \ref{fig:Teaser}, and consolidates them into a structured evaluation framework (Fig. \ref{fig:Framework}). The core of the framework defines four fundamental capability dimensions --- semantic, spatial, temporal, and physical --- that represent unique facets of scenario understanding. In addition, the framework introduces anticipation as the interrelationship between these dimensions to predict future developments and evaluate possible actions (Fig. \ref{fig:Teaser}). Further systematisation into context layers, input modalities, the four capability dimensions of scenario understanding, and downstream tasks lays the foundation for structured benchmarking and dataset creation.

By enabling a holistic evaluation of MLLMs in autonomous driving, this contribution supports future research on integrating MLLMs into interactive, language-guided autonomous systems. Ultimately, the framework aims to facilitate the development and systematic evaluation of generalisable and explainable foundation models for autonomous driving.

The key contributions of this paper are summarised as follows:
\begin{itemize}
    \item Derivation of key capabilities and core dimensions for MLLMs in autonomous driving applications.
    \item Proposing a holistic framework consolidating the existing literature for describing the required capabilities of MLLMs for comprehensive scenario understanding in autonomous driving.
    \item Framework formalisation outlining four core dimensions as a foundation for performance evaluation (semantic, spatial, temporal, physical).
    \item Exemplary application of the framework to two realistic, imaginary traffic scenarios to demonstrate its applicability.
\end{itemize}

\section{RELATED WORK}

\subsection{Capabilities of Human drivers} 
Earlier works outline the capabilities and requirements for human drivers to interact safely and efficiently in real-world traffic \cite{Abendroth2009}. The processing of a traffic scenario and the subsequent interaction are divided into the stages of perception, cognition and action. Accurate perception of the environment relies on the visual, acoustic, haptic and vestibular senses, with vision being the most dominant, providing up to 90\% of traffic-related information. The human eye processes visual input to identify objects, colours, motion, depth and size. In the cognition stage, this sensory information is combined with knowledge of traffic rules and an internal model of the environment, including physics like driving dynamics, to determine the most appropriate (safe and efficient) driving action.
The Responsibility Sensitive Safety (RSS) model \cite{shalev2017formal} applies the cognitive abilities of human drivers to make reasonable assumptions about worst case scenarios and formalise the \textit{Duty of Care} to the basic principles of autonomous driving systems. In this system an agent in the environment should never cause accidents or harm, should anticipate other drivers' misbehaviour by a \textit{useful driving policy}, and should obey the given traffic rules. This principle has been adapted in the Safety of the Intended Functionality (SOTIF) \cite{ISO.SOTIF.2022}, a standardised approach for evaluating the capabilities of autonomous driving systems in regard to safe functional behaviour.

\subsection{Requirements for Scenario Understanding} 
Scenarios are defined as the \textit{"[...] temporal development between several scenes in a sequence of scenes"}, where a scene represents a snapshot of the environment, including scenery, dynamic elements, and the self-representations of actors and observers, as well as their relationships to each other \cite{Ulbrich.TermsScenarioScene.2015}. The 6-Layer Model (6LM) \cite{Scholtes.6LayerModel.2021} provides a structured framework for describing driving scenarios in six different layers. Layer 1 focuses on the road network and its regulations, such as road markings and traffic signs. Layer 2 includes roadside structures, while Layer 3 addresses temporary changes to Layers 1 and 2, such as construction signs. Layer 4 contains dynamic objects, including vehicles and pedestrians, which are described in a time-dependent manner. Layer 5 covers daytime and weather conditions, and Layer 6 includes environmental factors and digital communication information.
In \cite{marti2019review}, the authors draw relationships between information and behavioural capabilities for autonomous driving. Based on these relationships, vehicle sensors for autonomous driving and the types of information they need to process for perception tasks are evaluated. These types of information include spatial configuration, kinematics, identification, regulation, and context, which serve as the basis for safe and efficient interaction with the environment. 
Building on these requirements, \cite{karle2022scenario} structures scenario understanding into pattern-based, motion-based, and physics-based approaches, and proposes different neural network architectures to capture the respective tasks within the spatial, temporal, and relational domains.

\begin{figure*}[t!]
    \centering
    \includegraphics{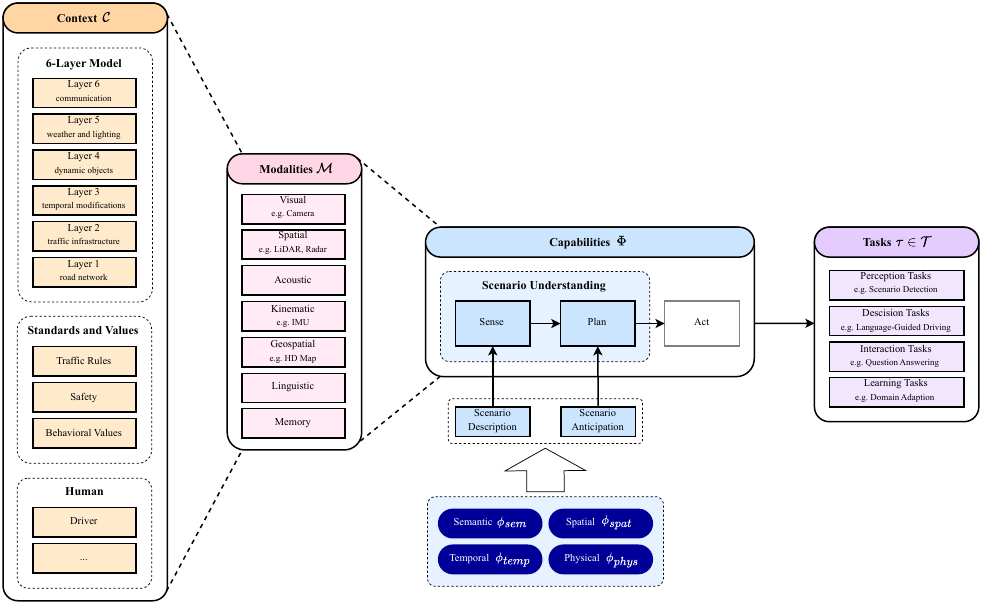}
    \caption{Framework for a capability-driven assessment of MLLMs' scenario understanding. The environmental context leads via perceptual modalities to key capabilities of scenario description and anticipation, with underlying capability dimensions along the sense-plan-act chain, resulting in executable tasks.}
    \label{fig:Framework}
\end{figure*}

\subsection{Foundation Models for Autonomous Driving} 
Recent research has introduced several models that use large language models (LLMs) for scenario understanding in autonomous driving tasks in the frame of proof-of-concepts. Models focusing on motion planning and forecasting highlight the potential of LLMs as decision makers, partially addressing spatial and temporal factors that are critical for effective scenario understanding, planning and action \cite{Seff_2023_ICCV, mao2023gptdriverlearningdrivegpt, wang2024hedrivehumanlikeendtoenddriving, sha2023languagempclargelanguagemodels, Huang.LLMasPlanner.2025}. Due to the image processing capabilities of current state-of-the-art MLLMs, models mostly interpret camera information either in sequences or in single snapshots of the scenario, which limits the effective integration of spatial and temporal information besides semantic aspects \cite{Huang.AlignPerceptionWithLLM.NEURIPS2023,tian2024drivevlm,Xu.DriveGPT4.2024,Zhang_2021_CVPR,jiao2025lavidadrivevisiontextinteraction}. In order to improve spatial capabilities, \cite{tian2024drivevlm} proposes a model to address difficulties in spatial and motion understanding within vision-language models (VLMs) by injecting explicit outputs from a 3D object detector in the context. In addition to semantic, spatial and temporal aspects, \cite{Fu.DriveLikeHuman.2024} further outlines three required high-level capabilities for VLMs in driving applications: reasoning, interpretation and memorisation. An approach to embodied scene understanding is proposed in \cite{Zhou_2025}, where VLMs are evaluated in terms of description, localisation, memory and prediction. As \cite{Zhou_2025} is focusing on question answering and model training, the requirements for scenario understanding capabilities are fragmented and lack a formal definition.
Recent surveys \cite{cui2024survey, yang2023llm4drive, zhou2024vision} outline the developments and applications of MLLMs in autonomous driving, focusing on the different models and consolidating primary tasks for these models mainly consisting of question answering and planning.

\subsection{Benchmarks and Datasets} 
For testing models towards scenario understanding, various datasets have been developed to enhance autonomous driving models, focusing on semantic, and partly on spatial, and temporal understanding. The OmniDrive-nuScenes benchmark \cite{wang2024omnidriveholisticllmagentframework} integrates semantic, spatial, and long-horizon reasoning, offering a holistic evaluation framework for autonomous driving models, while \cite{guo2024drivemllm} emphasise the importance of spatial reasoning for MLLMs in the domain of autonomous driving, providing an open-loop evaluation for different general aspects of spatial reasoning such as left/right, front/behind judgement and relative distances.
Furthermore, the NuInstruct dataset \cite{Ding_2024_CVPR} addresses a gap in existing research by providing multi-view and temporal information, as well as the proposed Rank2Tell dataset \cite{Sachdeva_2024_WACV}, which is focused on ranking the importance of objects and providing annotations of semantic, spatial, temporal, and relational attributes for scene understanding. DriveLM-nuScenes \cite{Chonghao.DriveLM.2024} focuses on integrating temporal aspects of object states, emphasising the dynamic nature of perception, prediction, planning, behaviour, and motion.
In contrast, most existing datasets predominantly assess semantic understanding, while spatial and temporal aspects are often underrepresented \cite{Kim_2018_ECCV_BDD-X,NuScenesQA_2024,Cao_2024_CVPR_MapLM,Xu.DriveGPT4.2024}.
The ability to understand physical models of the world is neglected by most approaches, although it is considered a fundamental aspect of common sense reasoning in traffic scenarios, while recent benchmarks highlight the limitations of MLLMs in this domain \cite{wang2023newton, motamed2025generative}.

\section{CAPABILITY-DRIVEN FRAMEWORK FOR SCENARIO UNDERSTANDING}

Based on the fragmented aspects identified in the related work, this paper proposes a structured framework, as illustrated in Figure \ref{fig:Framework}. It consolidates key aspects for deriving requirements on MLLMs for scenario understanding in autonomous driving. The framework streamlines the contextual information that defines the environment in which the agent interacts with other elements and agents. The context is captured by different modalities, represented by the sensors of the autonomous system, which serve as input for the scenario understanding process. This process includes the core capabilities of scenario description and anticipation along the sense-plan-act sequence.

These capabilities can be categorised into four core dimensions: \textbf{semantic} (what), \textbf{spatial} (where), \textbf{temporal} (when), and \textbf{physical} (how). Based on the MLLM’s capabilities, concrete tasks can be derived for the action phase, meeting the outlined requirements. Tasks can be classified into perception tasks, decision tasks, interaction tasks, and learning tasks.

\begin{figure*}[t!]
    \centering
    \includegraphics{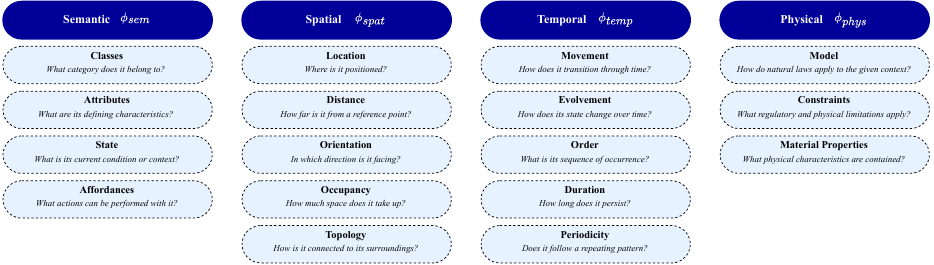}
    \caption{Four core dimensions of the descriptive capability of MLLMs. The dimensions are separated for explicit evaluation, facilitating targeted improvements of models.}
    \label{fig:Dimensions}
\end{figure*}

\subsection{Context} 
The context $\mathcal{C}=\{{C}_{\mathrm{6LM}},{C}_r,{C}_h\}$ structures the domain of the environment in which the autonomous agent interacts with elements of the environment and other agents. It consist of situative elements ${C}_{6LM}$ of a scenario as described by the 6LM \cite{Scholtes.6LayerModel.2021} as well as common standards and values ${C}_r$, such as traffic rules, safety considerations, and behavioural values, analogous to the RSS model \cite{shalev2017formal}. In addition to the environmental context, the driver ${C}_h$ must be considered for observing the system and interacting with the system in natural language by requesting information and giving instructions or triggering driver takeovers. As a consequence, the context captures all static and dynamic elements, as well as the general rules and required knowledge of the environment. In an ideal case, the agent needs to understand the context as complete as possible in order to derive the optimal action related to its task.

\subsection{Modalities} 
Modalities $\mathcal{M}$ are represented by the information available to the autonomous agent, captured as multimodal input data streams from vehicle sensors, for processing of the context with the aim to cover as much information as possible needed for the related downstream tasks. The sensory values  ${x}$ are expressed as a function of time $t$, formulated as: $\mathcal{M}={ \sum_{i \in \mathcal{I}}m_i(t,{x}) }$, where $\mathcal{I}$ represents the set of all modality types, and each $m_i(t,{x})$ represents the sensory input at time $t$ for a specific modality such as visual information from cameras ${m}_{\mathrm{vis}}$, spatial information capturing with Radar and LiDAR ${m}_{\mathrm{spat}}$, and acoustic information ${m}_{\mathrm{ac}}$. Kinematics ${m}_{kin}$ can be retrieved from kinematic sensors such as the Inertial Measurement Unit (IMU), that can aid in identifying the spatiotemporal state of the ego vehicle as well as capturing motion of other dynamic elements in the scenario. This is supported by the availability of geospatial information ${m}_{\mathrm{geo}}$ spanning from GPS positions to detailed HD maps containing abstracted representations of the broader environment. The questions and instructions by the driver can be captured by language ${m}_{\mathrm{lingu}}$ through speech-to-text interfaces. In addition, memorisation ${m}_{\mathrm{mem}}$ provides historical context by retaining information about past observations, decisions, and actions which is especially relevant for learning tasks. The modalities can be considered as the input data for MLLMs which needs to be processed in order to optimise the required capabilities of MLLMs in the related dimensions.

\subsection{Capabilities} 
Modalities and capabilities are inherently interconnected, as the quality of the perceived environment directly influences the maximum achievable quality with which the MLLM can describe the context across the four capability dimensions.
Along the sense-plan-act chain the key capabilities for scenario understanding start with scenario description, aiming to capture the context as complete as possible for subsequent anticipation through reasoning. 
This description process needs to capture the four underlying capability dimensions, as seen in Figure \ref{fig:Dimensions}.

\textbf{Semantic:}
The identification and description of different elements $e_i$ in traffic scenarios forms the semantic dimension $\phi_{\mathrm{sem}}$ of MLLM capabilities. Given the element class $K_i$  (e.g., vehicle, pedestrian), its corresponding attributes $A_i$ (e.g., color, type), the elements state $Z_i$ (e.g., parked, moving), and its affordance $\alpha_i$ (e.g., can occlude, can be run over), the set of semantic capability is defined as:
\begin{equation*}
    \begin{aligned}
        \phi_{\mathrm{sem}}(e_i)=\{ & (K_i,A_i,Z_i,\alpha_i) | \\ & K_i \in \mathcal{K},A_i \in \mathcal{A}, Z_i \in \mathcal{Z},\alpha_i \in \Lambda \},
    \end{aligned}
\end{equation*}
where $\mathcal{K}$, $\mathcal{A}$, $\mathcal{Z}$, and $\Lambda$ represent the corresponding sets of element classes, attributes, states, and affordances, respectively.

\textbf{Spatial:}
The spatial dimension $\phi_{\mathrm{spat}}$ describes the geometric properties of a scenario, which include the position ${p}_i$, distances to reference elements ${d}_i$, and elements orientations ${O}_i$ represented as rotation matrix. The occupancy $\mathrm{{occ}}_i$ describes the spatial extent of the element represented by its bounding volume (e.g., bounding box size $l, w, h$), while the topology $\mathrm{{top}}_i$ is a set of topological connections describing the spatial relation to other elements $\mathrm{{top}}_i=\{  (e_i,e_j,\mathrm{rel}_{ij}) | (e_j \in E, \mathrm{rel}_{ij} \in \mathcal{R}) \}$, with related elements $e_j$, the topological relation ${\mathrm{rel}}_{ij}$ between $e_i$ and $e_j$ (e.g., adjacency, containment, connectivity), and the set of possible relations $\mathcal{R}$ (e.g., left of, right of, above, touching). This relation results to:
\begin{equation*}
    \begin{aligned}
        \phi_{\mathrm{spat}} = & \{ (e_i,{p}_i,{O}_i,d_i,\mathrm{{occ}}_i,\mathrm{{top}}_i | {p}_i \in \mathbb{R}^3, \\
        & {O}_i \in SO(3),d_i \in \mathbb{R},\mathrm{{occ}}_i \in \mathbb{R}^3, \mathrm{{top}}_i \in \mathbb{T} \}.
    \end{aligned}
\end{equation*}
The resolution at which the spatial elements need to be understood depends on the downstream task, ranging from general descriptors to detailed relative distances of relevant entities in the traffic scenario. 

\textbf{Temporal:}
The temporal dimension $\phi_{\mathrm{temp}}$ represents the evolution of elements over a time duration of $T_i$, which include the velocity and acceleration vectors $v_i(t)$ and $a_i(t)$ describing movement. The temporal evolvement $s_i(t)$ of a scenario is the state of the element at time $t$, with $s_i(t) \in \mathcal{S}$, where $\mathcal{S}$ is the set of possible states (e.g., parked, moving, stopped). The temporal sequence of element occurrences is the order $\pi_i(t)$ with $\pi_i(t) \in \Pi$, where $\Pi$ represents the set of possible orderings (e.g., before, after, simultaneous). The periodicity can be derived from $s_i(t)$ by detecting recurring patterns over time. These aspects lead to:
\begin{equation*}
    \begin{aligned}
        \phi_{\mathrm{temp}} = & \{ (e_i,T_i,v_i(t),a_i(t),s_i(t),\pi_i(t)) | T_i \in \mathbb{R}, \\
        & v_i \in \mathbb{R}^3,a_i \in \mathbb{R}^3,s_i(t) \in \mathcal{S},\pi_i(t) \in \Pi \}.
    \end{aligned}
\end{equation*}
The temporal dimension can be either active behaviour of dynamic traffic participants or passively induced influences from other actors and the environment.

\textbf{Physical:}
The physical world model, respectively the rules applied for the semantic, spatial, and temporal dimensions $\phi_{\mathrm{sem}}$, $\phi_{\mathrm{spat}}$, and $\phi_{\mathrm{temp}}$, are defined within the physical dimension $\phi_{\mathrm{phys}}$. The physical dimension acts as a descriptive function that defines the physical models ${f}_{\mathrm{phys}}$ and constraints $\mathbb{C}$ governing the relations between the other three dimensions:
\begin{equation*}
    \phi_{\mathrm{phys}} = {f}_{\mathrm{phys}}(\phi_{\mathrm{sem}}, \phi_{\mathrm{spat}}, \phi_{\mathrm{temp}}), \quad s.t. \ \mathbb{C}.
\end{equation*}
The physical model includes various aspects including dynamic behaviour and physical properties (e.g., optical characteristics), while the constraints include both regulatory (e.g., traffic rules) and physical limitations.

As a result, the four core dimensions $\phi_\text{core}$ of the capability framework are semantic, spatial, temporal and physical:
\begin{equation*}         
     \phi_\text{core} = \{\phi_{\mathrm{sem}},\phi_{\mathrm{spat}},\phi_{\mathrm{temp}},\phi_{\mathrm{\mathrm{phys}}} \}.
\end{equation*}
The semantic, spatial and temporal dimensions have the property of disjunction, that is:
\begin{equation*}
    \phi_{\mathrm{sem}} \cap \phi_{\mathrm{spat}} \cap \phi_{\mathrm{temp}} = \emptyset.
\end{equation*}
All four dimensions describe different aspects of traffic scenarios over time. The temporal and physical dimensions inherently evolve continuously over the entire time interval $[T_\mathrm{s}, T_\mathrm{e}]$, representing how elements transition through time and obey physical and regulatory laws. In contrast, the semantic and spatial dimensions describe the state of the scenario at discrete time instances $t_i$, reflecting perceptual snapshots at specific moments. This distinction is formalised as:
\begin{equation*}
\begin{aligned}
&\phi_{\mathrm{temp}}, \phi_{\mathrm{phys}} : [T_\mathrm{s},T_\mathrm{e}] \implies \text{continuous },\\
&\phi_{\mathrm{sem}}, \phi_{\mathrm{spat}} : { t_1, t_2, \dots, t_N }, \quad t_i \in [T_\mathrm{s},T_\mathrm{e}] \implies \text{discrete}.
\end{aligned}
\end{equation*}
Subsequently, the \textbf{Scenario Description} $\Upphi_{\mathrm{D}}$ is the ability to describe a scenario through a context $\mathcal{C}$ mapped by a set of sensor modalities $\mathcal{M}$ to the four outlined core dimensions and to establish intercorrelations using a description function ${f}_\text{desc}$ from a starting point in past time $T_\mathrm{s}$ to the present, resulting to:
\begin{equation*}
      \Upphi_{\mathrm{D}}(T_\mathrm{s}) =  {f}_\text{desc}(\phi_\text{core},t),\quad t\in[-T_\mathrm{s},0].
\end{equation*}
Similarly, the \textbf{Scenario Anticipation} $\Upphi_{\mathrm{A}}$ is the ability to estimate possible future developments of forthcoming scenarios up to time  $T_\mathrm{e}$ given a prediction function ${f}_\text{pred}$ and the current scenario description $\Upphi_{\mathrm{D}}$:
\begin{equation*}
      \Upphi_{\mathrm{A}}(T_\mathrm{e}) = \Upphi_{\mathrm{D}}(T_\mathrm{s}) +  {f}_\text{pred}( \phi_{\mathrm{core}},t), \quad t\in(0,T_\mathrm{e}].
\end{equation*}
Consequently, the ability to describe and anticipate scenarios ultimately leads to \textbf{Scenario Understanding} $\Upphi_\mathrm{U}$:
\begin{equation*}
    \Upphi_\mathrm{U} = \Upphi_{\mathrm{D}} \cup \Upphi_{\mathrm{A}}.
\end{equation*}

Following scenario understanding, the selection of the right follow-up action $\mathcal{A}_i$ depending on the task is formalised by $\mathcal{A}_i = {f}_{\mathrm{task}}(\Upphi(e_i), \tau)$, where ${f}_\mathrm{task}$ is the task-dependent decision function, and $\tau$ represents the task specification. As a result, the best possible action $\mathcal{A}_i^*$ (e.g., answer to driver questions or language-guided driving) is derived.

Finally, the context, the modalities and capabilities all streamline into the set of tasks $\mathcal{T}=\{ \tau_\mathrm{p},\tau_\mathrm{d},\tau_\mathrm{i},\tau_\mathrm{l} \}$ which need to be carried out by the MLLM with its underlying scenario understanding. The set of tasks include perception tasks $\tau_\mathrm{p}$, decision tasks $\tau_\mathrm{d}$, interaction tasks $\tau_\mathrm{i}$, and learning tasks $\tau_\mathrm{l}$.

In general, the scenario understanding process of MLLMs can be defined as a function $\Upphi_\mathrm{U}$ mapping the context $\mathcal{C}$, modalities $\mathcal{M}$, and the MLLMs' capabilities $\Upphi$ to a task $\tau$ in a set of tasks $\mathcal{T}$, and deriving a best possible action $\mathcal{A}^*$:
\begin{equation*}
    \Upphi_\mathrm{U}: (\mathcal{C},\mathcal{M},\Upphi) \rightarrow \tau \in \mathcal{T},\mathcal{A}^*.
\end{equation*}

\section{EVALUATION WITH EXEMPLARY SCENARIOS}

Based on the proposed framework, two exemplary traffic scenarios at an imaginary urban intersection \cite{inDdataset} are presented to illustrate the capability dimensions in practice. The examples emphasise selected key aspects to provide a clear and concise demonstration of the framework's applicability. In a comprehensive application, all relevant information (e.g., all layers of the 6LM) must be incorporated to ensure a holistic scenario understanding. Furthermore, MLLMs should generate structured outputs that systematically cover all capability dimensions, aligning with the representation shown in the corresponding figures.

\subsection{Scenario 1: Taxi pick-up on the roadside}

\begin{figure}[t!]
    \centering
    \includegraphics[width=\linewidth]{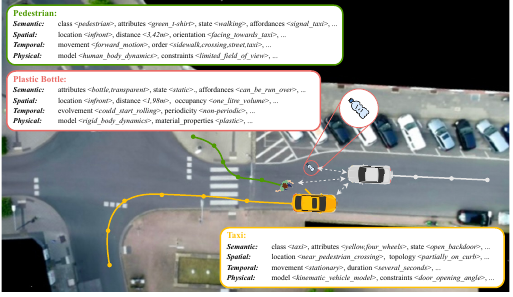}
    \caption{\textit{Scenario 1} shows a taxi pick-up situation at the roadside with a pedestrian crossing towards the taxi. A plastic bottle is lying in front of the ego vehicle (grey).}
    \label{fig:Scene1}
\end{figure}

\textbf{Context}.
Figure \ref{fig:Scene1} illustrates an urban intersection scene with multiple agents interacting. A pedestrian approaches a taxi for pick-up at the roadside while the ego vehicle nears the intersection. The pedestrian's crossing intention introduces a critical interaction, requiring the ego vehicle to yield. Additionally, a plastic bottle lying on the street acts as a secondary obstacle that could influence the driving decision. The combination of static and dynamic elements with potential occlusions makes this scenario suitable for evaluating multimodal scenario understanding.

\textbf{Modalities}.
A multimodal sensor setup equips the ego vehicle, including six cameras (front, front left, front right, back, back left, back right) offering full 360° perceptual coverage (visual), four corner radars providing distance and relative position measurements (spatial), a GPS sensor for global localisation (geospatial), and an IMU delivering motion dynamics information (kinematic). These modalities form the perceptual foundation for the ego vehicle’s understanding of the environment.

\textbf{Semantic Capability}.
Semantic understanding involves recognising relevant objects and their properties. The $<$\textit{pedestrian}$>$ represents a vulnerable road user with distinctive attributes such as a $<$\textit{green t-shirt}$>$. His state is classified as $<$\textit{walking}$>$, indicating ongoing motion. The pedestrian’s affordance implies that he $<$\textit{can signal a taxi}$>$ and $<$\textit{enter the vehicle}$>$, affecting the interaction dynamics. The $<$\textit{taxi}$>$ appears as a $<$\textit{yellow}$>$ $<$\textit{vehicle}$>$ with the $<$\textit{back door open}$>$, increasing the complexity of the scene by altering its spatial extent. Categorised as a $<$\textit{rigid}$>$ object with $<$\textit{transparent}$>$ material properties and $<$\textit{static}$>$ state, the $<$\textit{bottle}$>$ suggests an affordance that it $<$\textit{can be run over}$>$ without causing harm.

\textbf{Spatial Capability}.
Spatial capabilities enable MLLMs to estimate object locations, distances, and orientations. Positioned $<$\textit{in front}$>$ of the ego vehicle at approximately $<$\textit{3.42 meters}$>$, the pedestrian $<$\textit{faces towards the taxi}$>$. The taxi occupies a spot $<$\textit{near the pedestrian crossing}$>$ and $<$\textit{partially on the curb}$>$, impacting the available free space. A plastic bottle, detected $<$\textit{1.98 meters}$>$ ahead, occupies a small spatial extent corresponding to $<$\textit{one litre volume}$>$ with minimal influence on the drivable area.

\textbf{Temporal Capability}.
Temporal understanding enables predicting the dynamic evolution of objects. The pedestrian’s movement exhibits $<$\textit{forward motion}$>$, likely continuing towards the taxi as its temporal sequence follows $<$\textit{sidewalk → crossing street → entering taxi}$>$. The taxi is in a $<$\textit{stationary}$>$ state and expected to remain so for $<$\textit{several seconds}$>$ due to the ongoing pick-up process, while the plastic bottle maintains a static state over time. Yet its temporal evolution indicates that it $<$\textit{could start rolling}$>$ due to external forces such as wind or the movement of other vehicles.

\textbf{Physical Capability}.
Physical understanding considers objects' material properties and interaction potential. The pedestrian’s $<$\textit{human body dynamics}$>$ imply a $<$\textit{limited field of view}$>$, particularly towards the ego vehicle, necessitating a cautious approach. The taxi follows a $<$\textit{kinematic vehicle model}$>$, with the open backdoor increasing its spatial footprint and posing a potential collision risk. The plastic bottle, modelled by $<$\textit{rigid body dynamics}$>$, features $<$\textit{plastic}$>$ material properties making it non-harmful if run over by the ego vehicle.

\textbf{Tasks}.
Yielding to the pedestrian until he enters the taxi and the backdoor is closed represents the primary decision task. The decision relies on the pedestrian's semantic affordance and spatial position, aligned with traffic rules and safety standards. Once the pedestrian is no longer in motion and the taxi door is closed, the ego vehicle can proceed slowly. The plastic bottle is evaluated as a non-hazardous object that can be run over without risk.

To perform scenario understanding in this situation, an MLLM must identify object classes, attributes, and affordances (semantic), localise objects and estimate their relations (spatial), predict motion patterns and temporal evolvements (temporal) as well as anticipating their future intention, and assess object dynamics and operational constraints (physical). These capabilities enable a holistic understanding of the driving scene, forming the basis for safe and context-aware autonomous driving. 

\subsection{Scenario 2: Dynamic occlusion of cyclist about to pass a crosswalk}

\begin{figure}[t!]
    \centering
    \includegraphics[width=\linewidth]{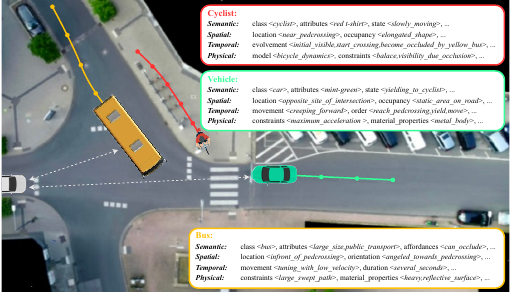}
    \caption{\textit{Scenario 2} shows a cyclist approaching a pedestrian crossing. The cyclist is occluded by a turning yellow bus from the perspective of the ego vehicles (grey).}
    \label{fig:Scene2}
\end{figure}

\textbf{Context}.
Figure \ref{fig:Scene2} depicts a four-way intersection with a pedestrian crosswalk. The ego vehicle (grey) approaches the intersection, while a yellow bus turns left across the intersection. A cyclist slowly moves towards the crosswalk, intending to cross. Fully occluded by the turning bus in the presented time step, the cyclist creates an additional challenge. Another vehicle yields on the opposite side of the intersection, waiting for the cyclist to cross. The scenario takes place in an urban environment with an infrastructure conducive to walking and mixed traffic. Traffic rules such as yielding to vulnerable road users and maintaining safe distances apply.

\textbf{Modalities}.
The sensor setup integrates six cameras (visual), a top-LiDAR (spatial), GPS (geospatial), IMU (kinematic), and a speech-to-text interface (linguistic) into the ego vehicle. Beyond the previously mentioned modalities in \textit{Scenario 1}, the vehicle now possesses interaction capabilities through the speech-to-text interface.

\textbf{Semantic Capability}.
Semantic understanding identifies the cyclist as a dynamic object of class $<$\textit{cyclist}$>$, with the attribute $<$\textit{red t-shirt}$>$. The cyclist’s state is $<$\textit{slowly moving}$>$, indicating a cautious crossing approach. Categorised as $<$\textit{public transport}$>$ in colour $<$\textit{yellow}$>$, the $<$\textit{bus}$>$ possesses the affordance $<$\textit{can occlude}$>$, influencing the perception process. Recognised as a $<$\textit{car}$>$ in the state $<$\textit{yielding to cyclist}$>$, the $<$\textit{mint-green}$>$ vehicle implies adherence to traffic rules.

\textbf{Spatial Capability}.
Spatial understanding locates the cyclist $<$\textit{near the pedestrian crossing}$>$, with an $<$\textit{elongated shape}$>$ occupancy indicating the bicycle’s geometry. Positioned $<$\textit{in front}$>$ of the pedestrian crossing, the bus fully blocks the cyclist from the ego vehicle’s view. The mint-green vehicle is situated $<$\textit{on the opposite side of the intersection}$>$, waiting at the crosswalk with a $<$\textit{static area on the road}$>$. Distance and orientation estimations to each object rely on spatial relations.

\textbf{Temporal Capability}.
The cyclist's temporal sequence follows $<$\textit{initial visible → start crossing → become occluded by bus}$>$. While the cyclist approaches the crosswalk, the bus performs a $<$\textit{turning maneuver with low velocity}$>$, lasting $<$\textit{several seconds}$>$. The mint-green vehicle executes a $<$\textit{creeping forward}$>$ motion while waiting for the cyclist to cross. Dynamic changes in object states over time shape the development of the scenario.

\textbf{Physical Capability}.
Physical reasoning models the cyclist with $<$\textit{bicycle dynamics}$>$, including constraints on $<$\textit{balance}$>$ and $<$\textit{visibility due to occlusion}$>$. The $<$\textit{heavy}$>$ bus is constrained by a $<$\textit{large swept path}$>$ and features $<$\textit{reflective surfaces}$>$. The mint-green vehicle obeys a $<$\textit{kinematic vehicle model}$>$, with $<$\textit{maximum acceleration}$>$ limits and a $<$\textit{metal body}$>$ affecting collision severity.

\textbf{Tasks}.
Yielding at the intersection until the bus clears the cyclist’s occlusion constitutes the primary decision task. Estimating the cyclist’s intended motion and anticipating reappearance is essential for this decision. The mint-green vehicle’s yielding state serves as an additional cue supporting the cyclist’s priority (learning task). Additionally, the driver is informed via the human-machine interface (interaction task) about the detected occlusion by the bus and the anticipated cyclist crossing. The interface provides a warning that the cyclist is temporarily not visible but expected to reappear. Due to the indicators of the surrounding environment, the autonomous system is instructed to maintain the yielding position.

Scenario understanding in this situation requires an MLLM to infer the likely motion of occluded objects based on temporal sequences, contextualise yielding behaviour, and assess occlusions as dynamic spatial relations. Core capabilities include temporal prediction and adherence to traffic rules for vulnerable road users.

The evaluation of exemplary scenarios demonstrates how the proposed framework facilitates the structured, capability-driven assessment of MLLMs for autonomous driving. Each core dimension --- semantic, spatial, temporal, and physical --- can be assigned real physical values where applicable, enhancing the quantitative evaluation of capabilities. The proposed framework can be implemented in MLLMs through chain-of-thought reasoning \cite{Wei_NEURIPS_2022_ChainOfThought}, enabling the model to query and reason over individual dimensions systematically. This approach provides a comprehensive baseline for scenario understanding and decision-making in autonomous driving contexts. Additionally, the framework supports anticipation and downstream tasks such as decision tasks, which are particularly relevant for autonomous driving applications like language-guided driving. By explicitly guiding the model to generate structured outputs, the framework supports a more interpretable and transparent assessment of the multimodal reasoning abilities of MLLMs.

\section{Conclusion}
This paper introduces a holistic framework for describing the required capabilities of MLLMs for comprehensive scenario understanding in autonomous driving. The proposed framework systematically organises the capabilities along the sense-plan-act paradigm, focusing on the scenario understanding stages of description and anticipation.
Description is further divided into four distinct capability dimensions: semantic, spatial, temporal, and physical. Each dimension captures a specific descriptive aspect of the scene, enabling a detailed, multifaceted representation. The anticipation capability builds on these dimensions by establishing relationships between them, allowing the MLLM to predict future scenario evolutions and evaluate possible actions to accomplish downstream tasks, thereby optimising decision making in autonomous driving.

Future research can utilise the framework to develop comprehensive evaluation concepts, including benchmarks and datasets, that holistically assess the scenario understanding capabilities of MLLMs. In addition, the framework provides a foundation for improving model architectures and training methods tailored to autonomous driving scenario understanding. By enabling a structured evaluation and comparison of different model approaches, the framework contributes to the advancement of foundation models for autonomous driving and supports the development of reliable, performant, and explainable MLLMs for this domain.

\addtolength{\textheight}{-0.5cm}  


\bibliography{root}

\end{document}